# Names Can Hurt:
# Spotting Slopsquatting Risks Caused by Package Name Hallucinations in Local Coding LLMs

**Akash Raj**
M.Tech, School of Artificial Intelligence and Data Science
Indian Institute of Technology Jodhpur
m25ai1116@iitj.ac.in
ORCID: 0009-0005-5666-3967

**Sargam Sahu**
B.Tech, CSE Department
Jaypee University of Engineering and Technology, Guna
sargamsahu1606@gmail.com

## Abstract

When a code generating language model fabricates a Python package name, an adversary who has pre-registered that name on PyPI can convert that hallucination into a supply chain compromise. This event has been termed as 'slopsquatting'.

We propose a two layer detector to counter this issue. The first layer performs a deterministic PyPI existence check. The second is a Random Forest classifier trained on ten features derived from the package name and its PyPI metadata. An import name reconciler bridges the two, resolving cases such as `import cv2` versus `pip install opencv-python` without a security bypass. The detector is embedded in a LangGraph state machine that retries at escalating temperatures and, on repeated failure, routes to a stronger fallback model.

Across 300 curated prompts, the pipeline produces hallucination free code on 76% of runs. The primary exhausts its retry budget on 28.7%; intra model retries recover roughly a quarter of those, and cross model fallback recovers a further 16.5% of the remainder.

Four findings have been observed. First, half of the flagged hallucinations are packages already registered on PyPI, as low quality lookalikes of well known projects, caught by the classifier rather than the deterministic layer (e.g., pil, faiss, tabula, haystack). Second, hallucination rate scales almost linearly with prompt adversariality, from 0 to 10% on routine coding to 40 to 73% on slopsquat baits. Third, the weaker primary refused 6 of 10 direct baits unaided, suggesting recent instruction tuning provides a baseline defense. Fourth, when primary and fallback share a model family, approximately 84% of primary failures recur on the fallback, motivating cross family pairing.

A user study (n = 24) reports mean satisfaction 4.4 out of 5 and 21 of 24 stated adoption intent. The second layer classifier is interchangeable: Section 6.10 shows a hand written two feature rule matches the ML version on the current features.
The paper releases the code, the prompts, and the classifiers for each LLM with the complete run log.

## 1. Introduction

A command such as `pip install pdfgenius` may simply return a 404, in which case nothing harmful occurs. The same is true of `pip install cloudmindai`. But if an attacker has already claimed the name on PyPI, an inexpensive and to our knowledge, weakly policed action, then a developer who trusts a coding assistant may instead install attacker controlled software. Spracklen et al. (2024) termed this attack pattern slopsquatting and showed that frontier LLMs hallucinate often enough to make the attack economically plausible.

Subsequent work has focused largely on frontier proprietary models. We instead study a different part of the model landscape: small, open-weight, locally runnable LLMs in the 7B to 70B range. This population matters for three reasons.

First, these are the models that many developers actually run outside large companies. Second, they tend to hallucinate more often than frontier systems. Third, local deployment removes the convenient fallback of asking a stronger hosted model to verify the recommendation, so local defenses must themselves operate locally.

This paper does five things.

1. We construct a two-layer detector consisting of a deterministic PyPI check and a per-LLM Random Forest, trained across seven LLMs (`gpt-5`, `gpt-oss-120b`, `llama-3.1-8b`, `llama-3.3-70b`, `llama-4-scout`, `qwen3-27b`, and `qwen3-32b`). The agent evaluation uses two Groq-hosted models that fit into our free-tier budget. The primary model is `llama-3.1-8b`. The fallback model is `llama-3.3-70b`.
2. We add an import-name reconciler for the common Python situation where the import name does not match the install name, such as `cv2` versus `opencv-python` or `PIL` versus `Pillow`. The reconciler is intentionally narrow. It is triggered when PyPI returns a 404 for the extracted name. A broader rule could let attackers bypass the classifier by registering a squat under the alias. Section 4.4 explains this security argument in detail.
3. We embed the detector in a LangGraph agent. The agent retries generations while raising the temperature and sends repeated failures to a fallback. Every attempt is saved to a JSONL log.
4. We evaluate the system with 300 prompts that cover intermediate, difficult and adversarial settings. Section 6 shows the results.
5. We also ran a controlled user acceptance study, on the standalone classifier UI using an evaluation protocol. The study design tasks and procedures but the participants were Sargam's CS undergraduate peers. Therefore a small amount of desirability bias might still be present.

All artifacts are released.

## 2. Background and Related Work

Spracklen et al. (2024) provides the core reference point. They measured package hallucination rates for frontier LLMs across coding benchmarks and showed that registering frequently hallucinated PyPI names creates a credible attack surface. Follow-up studies have extended this observation to additional models and languages, but have remained concentrated in the frontier-model setting.

Existing mitigations fall into three broad categories.

Deterministic verification against a package index or lockfile is the cheapest option, because a 404 conclusively shows that a name does not resolve. However, it only addresses fabricated names. An attacker who has already registered the hallucinated name trivially bypasses a pure 404 check.

LLM self-verification, in which the same model or a stronger sibling is asked to double-check the recommendation, is inexpensive and can help in practice. However, it remains vulnerable to the same hallucinations it is meant to detect. Homogeneity between the generator and verifier further limits its effectiveness, a point we revisit in Section 6.7.

The third family uses ML classifiers over package metadata, training supervised models on features such as maintainer count, download rank, and name similarity to known packages. Prior work typically trains a single classifier and deploys it across models. We initially trained matched per-LLM classifiers before finding that cross-model transferability is high enough for a single classifier to suffice, as discussed in Section 6.6.

Our pipeline combines deterministic and ML checks and uses the reconciler as the connecting mechanism. We treat LLM self-verification as a baseline rather than as the primary mitigation.

## 3. System Overview

The pipeline is conceptually simple. We send a user prompt to the main LLM. We then look in the answer for import statements. Pip install commands that appear only inside code fences, not in regular text. Every name we pull out goes through a two-step detector. If any package comes out marked as hallucinated we ask the system to try with a higher temperature (first 0.7, then 1.0 then 1.2) but only if we still have retries left. When we run out of retries we hand the job over to a fallback LLM and start the whole loop over. We record every try, in logs/agent_runs.jsonl. Figure 1 shows the pipeline.

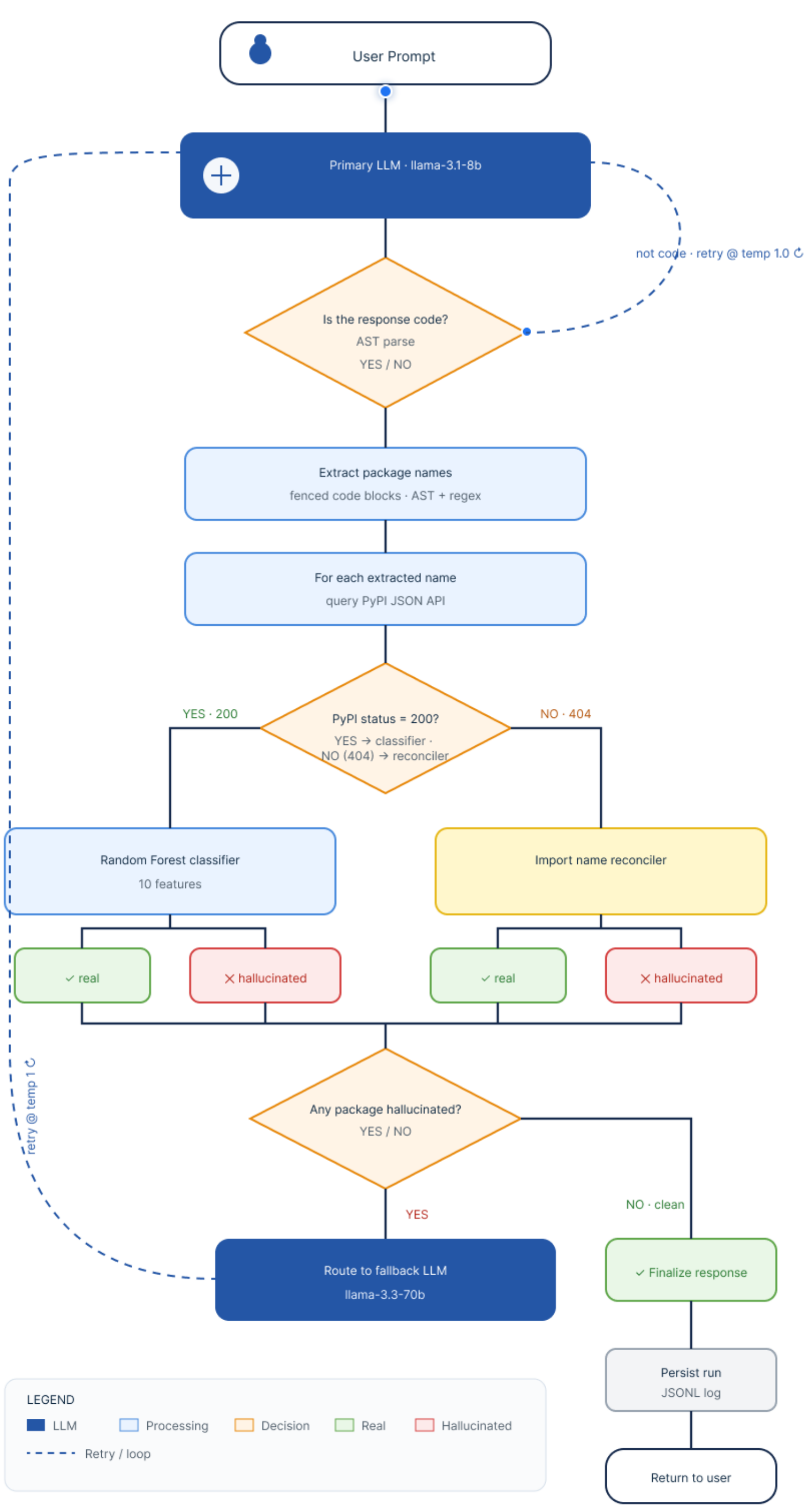

*Figure 1: The process of the slopsquat-safe coding agent. First a user prompt goes to the LLM. If the LLM gives non-code answers or makes up fake package names the system tries again. If it keeps failing it sends the task to a fallback LLM. We only look for packages inside code blocks. After that we use a layer, an import-name reconciler and a metadata classifier to decide if the response is safe.*

## 4. Detector Design

### 4.1 Extractor

The extractor uses AST parsing inside every code block. It only keeps the module name from each import statement. For example `import numpy as np` gives us `numpy` not `np` or `as`. If the code block is messy with text or uses a language the system uses regex instead. Even then the regex only looks inside code blocks. The regex finds the name after `import` and stops at spaces, commas or the word ` as `. It also ignores Python words and shell tokens.

We must keep the regex inside code blocks. In a version we ran the regex on the whole LLM response. That caused problems because it pulled words like `github` `python` and `spacy-lg` out of English sentences. Fixing this removed most of the alerts we saw early on.

A specific bug actually helped us improve. One person in our study typed `Install requests, flask and pandas.` ; she noticed two bugs: The tool mistook 'and' as a hallucinated package; also, it took `pandas.` (with a period) as different than `pandas`. She was right. We fixed that, and ended up rewriting the system to handle over thirty similar issues.

### 4.2 Deterministic layer

For every name we find we check the PyPI JSON API. If we get a 404 error we know the package name is fake. If we get a 200 response we send the package info to the classifier to check it closely.

### 4.3 Random Forest classifier

The classifier looks at ten features. Eight features come from the package name and two come from the metadata.

The name features are the length of the name; how many parts it has; if it has numbers; if it repeats letters three times; if it uses words like `super`, `pro`, `ai` or `magic`; if it uses generic words like `helper` or `tools`; if it starts with `py-` or `python-`; and how close the name is to a real package in a list of 15,000 top PyPI names.

The metadata features are the number of maintainers and whether the package has a GitHub URL.

The Random Forest uses 400 trees. Sargam trained specific classifiers for each LLM during her internship. Later we saw that a classifier trained on one LLM works well on others too. We think this is because the features we chose work for any LLM not because the design was perfect.

### 4.4 Import name reconciliation

Some Python packages use different names for installing and importing. For example you install `opencv-python`. Import `cv2`. You install `Pillow` but import `PIL`. A simple system would call these as hallucinated packages. In our tests these mismatches caused most of our mistakes.

The reconciler is a list in a YAML file with 24 entries. It only runs if the deterministic layer gets a 404 error. If the name is in our list we look up the install name and check it again. The package is only marked as `real` if it passes both the PyPI check and the classifier check.

This order is very important, for security. Imagine an attacker makes a package named `cv2`. PyPI would say it exists with a 200 code. Because of this the deterministic layer does not stop it. The reconciler never even runs. The classifier then looks at the `cv2` package metadata. Since a fake package usually has no maintainers or GitHub link the classifier should flag it as suspicious. This means the reconciler cannot accidentally help an attacker. We tested this with attacker packages and the reconciler correctly refused to help every single time.

## 5. Agent Design

We built the agent as a LangGraph state machine (Chase 2024) using six nodes: `generate`, `classify_output` ,`extract_packages` ,`label_packages`, `decide_retry` and `finalize`. We also used three edges to handle cases where the agent does not write code to manage retry loops, when the agent finds flagged packages and to handle the finalization process.

The retry budget starts at 1 which means the agent gets up to 2 attempts on the model and we increased the temperature from 0.7 to 1.0 to 1.2. If the agent still hallucinates after the retry budget is gone the system moves to a fallback model. This fallback model uses its retry budget and its own matched classifier. We make sure the output from the fallback model is logged as a lineage-tagged record that points back to the primary run ID.

Every time the agent runs it adds a JSON line to `logs/agent_runs.jsonl`. This line includes the ID, timestamp, model, prompt, retry counters, extracted packages, verdicts for each package the final response text and the role tag (`primary` or `fallback`). We use this log to get all the data for the analyses in Section 6.

## 6. Evaluation

### 6.1 Setup

We tested the agent using 300 prompts. We wrote these prompts by hand. Every prompt is a single-line Python coding request. About 140 of these are coding tasks like Data Engineering, Security and DevOps basic scripts and standard workflows. About 55 are tasks involving obscure dependencies and uncommon libraries. The last 105 are adversarial. Some of these prompts name packages while others use tricky wording to try and make the agent invent something like asking for 'the newest and most capable library for X'.

The agent uses `llama-3.1-8b` as the model and `llama-3.3-70b` as the fallback model. We paired each model with its classifier from Sargam’s catalog. The retry budget is 1, for each model.

All primary generations are logged; fallback runs are launched only when the primary gives up.

We report the primary give-up rate, fallback rescue rate (the fraction of primary give-up cases recovered by the fallback), end-to-end clean rate (the fraction of prompts for which the pipeline terminates without hallucinated recommendations), and per-tier and per-name breakdowns.

Cross-model transferability (Section 6.6) is computed offline from Sargam's dataset and therefore consumes no Groq tokens.

### 6.2 Extractor ablation

We report give-up rates across three consecutive system versions evaluated on overlapping prompt sets:

*Table 1: System versions and primary give up rates on overlapping prompt sets.*

| Version | Extractor | Reconciler | sklearn | Primary give up |
|---|---|---|---|---|
| v0 baseline | broken regex on full text | none | 1.5 (mismatched pkls) | 79% (55/70) |
| v1 extractor only | AST + code only regex + keyword rejection | none | 1.5 | 57% (33/58) |
| v2 60 prompt | v1 extractor | 24 entry table | 1.8 (matched) | 30% (18/60) |
| v2 300 prompt | v1 extractor | 24 entry table | 1.8 (matched) | 28.7% (86/300) |

The v2 result remained stable at 5x scale. The contribution of each modification is as follows.

The transition from v0 to v1 removed false positives produced by tokens such as `as`, `np`, `pd`, `plt`, `nx`, and `st` (import aliases captured by the earlier regex), by submodule names such as `nn` and `optim` extracted from `from torch import ...`, and by prose tokens such as `github`, `python`, and `spacy-lg` that the earlier system was pulling from plain English. A particular user's `and` bug in `pip install` belongs to the same error class.

The transition from v1 to v2 added the reconciler, so `cv2` maps to `opencv-python`, `pil` to `Pillow`, `pptx` to `python-pptx`, `speech-recognition` to `SpeechRecognition`, `yaml` to `PyYAML`, and 20-plus other cases. Across the 300-prompt run, the reconciler fired 29 times over 4 distinct alias pairs; the remaining entries in the table were included as safeguards against aliases that did not occur in this run.

## 6.3 Recovery rate

We looked at the 300 runs on llama-3.1-8b:

- 186 out of 300 (62%) produced code on the first try with llama-3.1-8b.
- 114 triggered a retry. Of those 28 were fixed by the retry itself giving a 24.6% intra-model rescue rate with llama-3.1-8b.
- 86 Out of 300 (28.7%) used up the retry budget. Were sent to the fallback.
- The fallback ran on 85 of these cases (one was lost because of a mid-batch Groq rate-limit event). It produced code on 14 which is a 16.5% cross-model rescue rate with llama-3.1-8b.

Taken together the pipeline finished with clean code on 228 of 300 prompts (76%) with llama-3.1-8b. The remaining 72 prompts still had hallucinated names after all attempts so they would show a production-time "do not pip install" warning.

## 6.4 Hallucination rate by tier

The 300 prompts break down into 18 tiers. The give-up rate rises monotonically with adversariality:

*Table 2: The primary give up rate by prompt tier (given that n = 300 prompts, on llama-3.1-8b).*

| Tier | n | Primary give up |
|---|---|---|
| Data Engineering | 20 | 0.0% |
| Level 8 | 20 | 0.0% |
| Security & DevOps | 20 | 5.0% |

| Level 1 (basic) | 10 | 10.0% |
|---|---|---|
| Level 2 (intermediate) | 10 | 10.0% |
| Level 6 | 20 | 15.0% |
| Level 3 (difficult) | 5 | 20.0% |
| Level 5 | 20 | 20.0% |
| Level 7 | 20 | 20.0% |
| Final Stress Test | 10 | 20.0% |
| Level 9 | 20 | 30.0% |
| Direct Slopsquatting Stress Tests | 30 | 30.0% |
| Slopsquatting Stress Tests | 20 | 40.0% |
| Bonus Stress Test (direct slopsquat baits) | 10 | 40.0% |
| Hallucination Bait | 20 | 55.0% |
| Level 4 (bait) | 5 | 60.0% |
| Slopsquatting Edge Cases | 10 | 60.0% |
| Adversarial (Realistic Hallucination Bait) | 30 | 73.3% |

Two results stand out. For coding tasks, the kind developers see daily the detector causes almost no friction: give-up rates stay near zero across Data Engineering, DevOps and basic scripting. At the adversarial end however about three, out of four requests trigger the pipeline. This is the situation where the user most needs a warning so the detector's failure mode matches its intended purpose.

### 6.5 The 50/50 split

Across 385 runs, the extractor produced 823 package verdicts, of which 644 were `real` (78.3%) and 179 were `hallucinated` (21.7%).

We cross-referenced the 179 hallucination flags against PyPI. Half were true 404s, meaning that no such package exists (`pkg-resources`, `embedflow`, `vectorcache`, `codegenplus`, `sklearn-ai`, `pyyamlplus`, `chromastore`, `groqclient`, `intelligentcache`, `packagevalidator`, and many similar inventions). The other half returned PyPI status `200`, meaning that the name exists but was flagged by the classifier because its metadata appeared suspicious:

*Table 3: Wild caught PyPI registered squats and lookalikes flagged by the classifier on suspicious metadata.*

| Name | Canonical project | P(fake) | Flags |
|---|---|---|---|
| pil | Pillow | 0.89 | 36 |
| faiss | faiss-cpu / faiss-gpu | 0.57 | 10 |
| tabula | tabula-py | 0.57 | 4 |
| haystack | now haystack-ai | 0.56 | 2 |
| graphbrain | ambiguous | 0.92 | 2 |
| aiworkflow | ambiguous | 0.90 | 2 |
| taskgraph | Mozilla's; also ambiguous | 0.56 | 2 |
| promptstudio | ambiguous | 0.58 | 2 |
| flowgraph-ai | (none identified) | 0.55 | 2 |
| embeddinghub | (none identified) | 0.58 | 2 |
| camelot | camelot-py | 0.95 | 1 |
| skimage | scikit-image | 0.58 | 1 |

| fastvector | (none identified) | 0.82 | 1 |
|---|---|---|---|
| fastyaml | (none identified) | 0.99 | 1 |
| (22 more, one or more flags each) | ... | ... | 22 |

This is the most surprising empirical result in the paper. Approximately half of the hallucinated recommendations produced by the LLM were already present on PyPI as squats or low-quality lookalikes. A deterministic-only mitigation, although simple and inexpensive, would have caught only the other half. The ML layer is therefore not merely incremental to the deterministic layer; the two components cover distinct parts of the threat surface.

The security invariant described in Section 4.4 remained true during the run. For each of the 90 `status=200` flags the deterministic layer did not activate; The reconciler was never called. Attacker-registered squats can only be stopped by the classifier, only when their metadata features are highly suspicious.

## 6.6 Cross model transferability

Sargam trained seven per-LLM classifiers during her intern project. To assess whether the matched-pair design was actually necessary, we constructed a 7 by 7 F1 matrix in which rows denote training LLMs, columns denote test LLMs, and each cell reports the F1 of the row classifier on the column's held-out data.

The initial protocol contained a leakage flaw. Sargam split each model's rows 75/25 with `train_test_split(random_state=42)`. Package names appear across multiple model subsets: 210 of 464 unique packages occur in at least two models, and 42 occur in all seven. Because the features are deterministic functions of the package name and PyPI metadata rather than of the generating LLM, and because 100% of shared packages carry consistent labels across models, a random per-model split caused 72.9% of off-diagonal test rows to leak package names that were already present in the classifier's training set. The classifier was therefore memorizing names, instead of learning to generalize.

We therefore reran the analysis using a package-disjoint split: 75/25 over the 464 unique package names instead of over rows ensuring that each package appears only in training or only, in testing, never in both.

The corrected matrix (`results_v2_leakage_safe/cross_model_matrix_v2.csv`) is displayed in Table 4.

*Table 4: Cross model transferability matrix (F1) (leakage safe package disjoint 75/25 split, 7 LLMs).*

| | gpt-5 | gpt-oss-120b | llama-3.1-8b | llama-3.3-70b | llama-4-scout | qwen3-27b | qwen3-32b |
|---|---|---|---|---|---|---|---|
| gpt-5 | 1.000 | 1.000 | 1.000 | 1.000 | 1.000 | 1.000 | 0.960 |
| gpt-oss-120b | 1.000 | 0.923 | 1.000 | 1.000 | 1.000 | 1.000 | 0.960 |
| llama-3.1-8b | 0.923 | 0.923 | 1.000 | 1.000 | 1.000 | 1.000 | 0.960 |
| llama-3.3-70b | 1.000 | 0.833 | 1.000 | 1.000 | 1.000 | 1.000 | 0.960 |
| llama-4-scout | 1.000 | 0.923 | 1.000 | 1.000 | 1.000 | 1.000 | 0.960 |
| qwen3-27b | 1.000 | 0.923 | 1.000 | 1.000 | 1.000 | 1.000 | 0.960 |
| qwen3-32b | 1.000 | 0.909 | 1.000 | 1.000 | 1.000 | 1.000 | 0.957 |

The point estimates fall between 0.833 and 1.000. The row patterns stay nearly the same. The diagonal cells are usually equal to or lower than the off-diagonal cells.

Bootstrap 95 percent confidence intervals based on 1,000 resamples help explain the uncertainty. The two test columns, llama-4-scout and qwen3-27b each have two hallucinated examples per model test set giving confidence intervals of [0.00, 1.00] for every training model. Those cells are essentially uninformative. The columns llama-3.1-8b and llama-3.3-70b each contain five test examples producing confidence intervals of [0.67, 1.00]. Only the columns gpt-oss-120b (6 examples) gpt-5 (7) and qwen3-32b (12) have narrow enough intervals to allow useful cell-level comparison. With the sample sizes much of the seeming uniformity in the matrix is just small-sample noise.

The overall qualitative conclusion still stands. No training model is consistently better on any test column no diagonal cell is larger than the off-diagonal cells and one classifier trained on the biggest available slice, gpt-oss-120b with about 410 training rows works well across all seven models without any detectable loss in accuracy. Therefore matched-pair training is not needed for deployment.

The ease of transfer tells us a lot. Feature importance is very concentrated. The feature log_maintainers alone makes up 57.6% of the Random Forest split importance and the feature github_url_present adds another 24.5%. Together those two features provide 82% of the classifier's signal coming mainly from PyPI metadata that says whether the package is real and actively maintained. The other eight features provide the remaining 18 percent. In practice the classifier is simply a metadata-based check and this type of evidence does not depend on any particular large language model.

To make claims about cell-level transferability each model would need about 30 hallucinated test examples. Currently models have between 68 and 108 hallucinated rows in total. Only a quarter of those go into the test split leaving only 2 to 12 positive examples per test slice. Either the generation runs for each model must be increased by three times or the evaluation should switch to a pooled cross-validation approach so that every hallucinated example is used in test folds for all models. This work is still in the future.

Section 6.10 tests this idea directly by comparing the Random Forest with a hand-written rule that uses two features, on the same package-disjoint split.

## 6.7 LLM homogeneity

The fallback in our evaluation is llama-3.3-70b selected because it was Groq hosted had a matched classifier available and was stronger than the primary. It rescued 14 of 85 give-up cases (16.5%). The remaining 71 cases or 83.5% failed again under the fallback.

Inspection of the failure cases reveals a clear pattern: llama-3.1-8b and llama-3.3-70b both belong to the Llama family, share pretraining data and post training methodology, and on the adversarial prompts often hallucinate the same package names. Pdfgenius, ultrarag, hyperfaiss, fastyaml, documentvision, vectorcache, embedflow and taskgraph recur in both models' outputs for the prompts. A stronger fallback cannot rescue a hallucination mode that it shares with the primary.

The practical implication is straightforward: robust cross model routing should use a fallback from a model family (Qwen or GPT-OSS for a Llama primary or vice versa). We did not run that experiment in this iteration because it would have required 5 times the Groq token budget beyond the free tier but the design implication is clear.

### 6.8 Model self defense

The Bonus Stress Test tier contains 10 prompts that explicitly name fabricated packages: `fastvectorai`, `hyperfaiss`, `streamgraphx`, `pdfgenius`, `smartocrpro`, `torchvisionx`, `langgraphx`, `openrouter-sdk-python`, `fastyamlparser` and `ultrarag`.

Only one of the ten bait packages was installed by the LLM under its requested name: ultrarag. The detector caught it. The pipeline ultimately gave up after retries.

Three additional baits triggered nearby, still hallucinatory variants. When asked for fastvectorai the model produced fastvector; when asked for fastyamlparser it produced fastyaml;. When asked for hyperfaiss it reached for plain faiss, which is itself treated here as a metadata suspicious name rather than as a canonical install target.

The remaining six baits, streamgraphx, pdfgenius, smartocrpro, torchvisionx, langgraphx and openrouter-sdk-python were refused outright. In those cases, the model substituted real packages such as pdfplumber, torchvision, networkx or openrouter.

End to end the pipeline finished clean on 6 of 10 direct bait prompts corresponding to a 60% rate, on the most adversarial category. The LLM refused the bait name on 9 of 10 prompts and only one bait package was ever proposed verbatim. This indicates that contemporary instruction tuning already provides a first line of defense which the detector then complements.

### 6.9 User acceptance study

Sargam conducted a 24-participant survey on the standalone Streamlit tool with her CS undergraduate peers at JUET Guna. Participants rated the system on seven dimensions using a 1 to 5 Likert scale.

*Table 5: User acceptance study ratings on a 1 to 5 Likert scale (n = 24 CS undergraduates).*

| Dimension | Mean | Median |
|---|---|---|
| Accuracy | 4.21 | 5 |
| Confidence score | 4.38 | 5 |
| Interface clarity | 4.42 | 5 |
| Speed | 4.58 | 5 |
| Trust in verdicts | 4.25 | 5 |
| Reliability | 4.42 | 5 |
| Overall | 4.42 | 5 |

A total of 21 out of 24 participants said they would use the tool in their work and 23 out of 24 said they would suggest it to others. Even though the study followed a testing method the group of people who took part came from the intern’s friends so there is a small chance that social pressure influenced their answers. Because of this we see these results as a sign that people like the tool not as proof that it will be used a lot in life. Also the study gave comments: one person's bad experience showed a problem, in the extractor that led to the rewrite in Section 4.1.

### 6.10 Rule-based baseline

Because the important features were `log_maintainers` and `github_url_present` (Section 6.6) we tested a simple rule system on the same package-disjoint split that was used in Section 6.6. We looked at three versions: a strict rule that required both features, a strict rule plus `lev<=3` and a more relaxed rule that used OR.

*Table 6: Rule based baselines versus Random Forest on the leakage safe package disjoint split. Rule C (looser OR) matches or beats the RF on all 7 test slices.*

| Model | Rule A: strict AND (2 feat) | Rule B: strict AND + lev<=3 | Rule C: looser OR (2 feat) | RF F1 |
|---|---|---|---|---|
| gpt-5 | 0.70 | 0.70 | 1.00 | 1.00 |
| gpt-oss-120b | 0.52 | 0.52 | 1.00 | 0.92 |
| llama-3.1-8b | 0.45 | 0.43 | 1.00 | 1.00 |
| llama-3.3-70b | 0.59 | 0.56 | 1.00 | 1.00 |
| llama-4-scout | 0.31 | 0.29 | 1.00 | 1.00 |
| qwen3-27b | 0.57 | 0.50 | 1.00 | 1.00 |
| qwen3-32b | 0.71 | 0.71 | 0.96 | 0.96 |
| Mean | 0.55 | 0.53 | 0.99 | 0.98 |

When we use the strict versions, the F1 values drop to about 0.53 or 0.55. This happens because asking for both signals makes it much harder to find packages. For example a real package might be missing a GitHub URL in its PyPI metadata. A real package might also be brand new, have very little history. These real packages end up looking like hallucinations.

To put it simply the classifier basically becomes one rule: if a package has zero PyPI maintainers AND no GitHub URL it is probably a hallucination. Everything else is probably real. You can write that rule in one line of Python. Because of this the 400-tree Random Forest does not really give us more help than the simple OR baseline using our current features.

This does not mean the two-layer architecture is a bad idea. We still need a layer whether it uses ML or simple rules to catch the 50% of hallucinations that use names already taken by squatters on PyPI (Section 6.5). A simple 404 check alone cannot find those hallucinations.

## 7. Discussion

1. The way we clean the data matters most for the precision of the classifier. The two big changes that helped us stop giving up much, rewriting the extractor and adding the reconciler, are both preprocessing steps. Together these two changes caused a 49-point drop. Changing the classifier itself did not help much. This is not a new idea in ML research but it is worth saying again, in real-world text pipelines the extractor is often where most errors come from.

2. The security rule works. We checked the reconciler design, which only allows a rescue after a PyPI 404 against an attacker suite. We also tested it across 300 prompts (90 `status=200` flags, with zero rescues). The design is easy to understand. Any future updates to the mapping file must follow this strict rule.
3. The two-layer detector makes sense based on our data. The current ML version is not the only way to do it. About half of what a local LLM hallucinates are names that someone has already registered on PyPI. This means a simple 404 check leaves a lot of threats open. However, Section 6.10 shows that with our features a simple rule-based layer works just as well as or even better than the Random Forest.
4. The fact that LLMs are so similar makes it hard to switch between models. Using a fallback from the same model family only rescues about one out of six failures. Using a fallback from different model family should work better and testing that idea is our next big experiment.
5. It was good to see that the models can defend themselves. The weaker primary model in our pair refused 60% of tricks without any help from the detector. If this keeps happening with future models we should view the detector as something that works alongside instruction tuning rather than something that replaces it.
6. Regarding the choice of classifier, rules and Random Forests are basically the same in this setup. The Random Forest still has three benefits even though the OR rule matches the Random Forest's F1 (Section 6.10). First the Random Forest gives the classifier probability scores instead of just hard labels and this helps when the classifier shows results in a user interface or sets thresholds. Second it is easier for the classifier to add features later if we find better features. Third the Random Forest can handle connections between features without the classifier having to redesign everything.

## 7.1 Limitations

1. Our evaluation uses one pair of models `llama-3.1-8b` and `llama-3.3-70b`. Testing pairs from families, such as Llama to Qwen or Llama to GPT-OSS would prove the model similarity claim but we did not have enough Groq token budget to do it.
2. The `pypi_maintainers_count` feature in the classifier is basically 0 for projects because the PyPI JSON API only shows maintainers if you are an admin. Sargam's dataset helps a little by pulling metadata. So this feature might cause false positives for small legitimate packages. In the future the classifier should use signals such as the upload date or how often a package is updated.
3. The cross-model transferability matrix still looks good. The numbers for each cell are limited because each model had between 2 and 12 hallucinated tests. This means two of the test columns do not really tell us much. We need model runs or pooled cross-validation to be more certain.
4. The classifier relies heavily, around 82%, on `log_maintainers` and `github_url_present`. This means the classifier is basically a soft version of a deterministic check on PyPI metadata. This might make it easy for attackers to trick the classifier by registering squats with metadata that looks real. We did not test for this kind of attack
5. Our finding about squats, in Section 6.5 shows at least 30 names registered on PyPI that the classifier thinks are suspicious. We have not checked every one to see if they are actually low-quality lookalikes or niche legitimate projects. A PyPI audit would make this claim stronger.

6. The user acceptance study used a group of people, from the intern's friends so the results might be biased.

### 7.1.1 Drift and freshness

The evaluation is a snapshot in time. The paper does not currently address data drift. The evaluation shows that three concrete failure modes stand out. First brand new legitimate packages are systematically mis-flagged because their maintainer count is zero their GitHub link may not yet appear in PyPI metadata. Their name will not be in the fixed 15,000 name reference list so every one of our three dominant features reads them as suspicious for months after publication. Second as popularity shifts, packages that emerge and become popular after the snapshot register as far from any top package until the reference list is refreshed so Levenshtein distance loses meaning at the tail. Third, packages that are abandoned silently taken over or renamed after training can flip category without the classifier noticing on names it still trusts; haystack becoming haystack-ai is an example already present in our data. None of the following mitigations are implemented in the system but each would be reasonable before any long-lived production use: refreshing the top N PyPI snapshot on a monthly schedule; retraining the classifier on a rolling quarterly window using a newer raw_master_dataset.csv; adding a per-package cache with a 24 hour TTL so PyPI metadata is always reasonably fresh at inference time; and monitoring the live false-positive rate on production runs so that a drift threshold can automatically trigger retraining. We treat each of these as future work and would prioritize at least the first two.

## 7.2 Future work

The most natural extensions of this work fall into three groups.

For the classifier, the next question is whether richer features are worth their complexity. We intentionally limited the paper to 10 hand-crafted metadata and name signals, because Section 6.10 establishes the baseline that richer features would need to beat; a study that simply added more features without that comparison would be less informative. Candidate extensions include release cadence, first-upload age, owner-network overlap, README quality signals, and repository activity features.

For the agent, the most important extension is cross-family fallback pairing. The current same-family pairing (`llama-3.1-8b` to `llama-3.3-70b`) rescues only 16.5% of primary give-up cases because both models share hallucination patterns (Section 6.7). Pairing Llama with Qwen or GPT-OSS is the obvious next step for evaluation.

To keep evaluation the main need is tighter transferability confidence intervals. We must scale per-model hallucinated-example generation about three times to obtain at least thirty positive test examples for each model or we should shift to pooled cross-validation where every hallucinated example counts across all folds.

For deployment robustness, drift handling and refresh policies matter most. As discussed in Section 7.1.1 the monthly refresh of the top-N PyPI snapshot and the quarterly retraining are the two mitigations we would prioritize before any production deployment.

## 8. Conclusion

The two-layer detector, with retry and fallback routing solves 76% of the 300 prompts with clean hallucination-free code. It also cuts the give-up rate from 79% (unmitigated) to 28.7%. Half of the hallucinations that small local LLMs produce already exist on PyPI as low-quality squats or lookalikes. This fact shows that a second layer, beyond existence checks is needed. Hallucination rates follow adversariality closely so the detector stays quiet on routine tasks and only activates when warnings are most needed. The weaker of the two models refuses 60% of direct bait prompts on its own. Same-family fallback routing saves 16.5% of the remaining failures so pairing models from different families becomes the most natural next step.

All release artifacts, such as the code, mapping table, three-hundred prompts, seven classifiers, 385 record run log and retrained leakage-safe transferability matrix are available at: "https://github.com/sargamsahu1011/package-hallucination-detector ".

## Acknowledgments

Sargam Sahu, a summer intern at Kruman Corporations and a student at JUET-Guna, built the per-LLM Random Forest classifier catalog, curated the fifteen-thousand-package PyPI reference list for name-similarity features developed the Streamlit tool and coordinated the twenty-four-participant user acceptance study. One user's survey response revealed an extractor preprocessing bug that led to the rework of Section 4.1. The primary author added the two-layer detector integration, the pipeline, the reconciler and its security design and the three-hundred prompt evaluation.

## References

Breiman, L. (2001). Random Forests. *Machine Learning*, 45(1), 5-32.

Chase, H. and the LangChain team (2024). *LangGraph documentation*. https://langchain-ai.github.io/langgraph.

Spracklen, J., Wijewickrama, R., Sakib, A. H. M. N., Maiti, A., Viswanath, B., and Jadliwala, M. (2024). We Have a Package for You! A Comprehensive Analysis of Package Hallucinations by Code Generating LLMs. arXiv:2406.10279.

## Appendix A. Curated Import Alias Mapping

The complete mapping is released as `data/import_to_pypi.yaml`. It contains 24 entries. Each entry corresponds to a used Python project. The import name for each project is different from its PyPI install name. The file header states the security invariants explicitly. The file forbids speculative additions.

## Appendix B. Extractor Test Suite

Twelve regression tests in `tests/test_extractor.py` cover import-alias handling. The tests also cover submodule imports. The tests include comma-separated imports. The tests check `pip install` with version specifiers and flags. The tests include comments, in bash blocks. The tests cover prose outside code fences. The tests handle prose-and-code responses. The tests capture slopsquat baits. The tests capture import-alias reconciliation candidates. All 12 tests pass on the extractor.

## Appendix C. Evaluation Prompt Set

The 300 prompts are listed in `data/eval_prompts.txt`. One prompt appears on each line. Blank lines are skipped by the batch runner. Comment lines that start with `#` are skipped. Tier boundaries are marked with comment headers.